\documentclass{article}
\usepackage{amssymb}
\usepackage{graphicx}
\usepackage{amsmath}
\usepackage[flushleft]{threeparttable}
\usepackage{makecell,booktabs}
\usepackage{caption}
\usepackage{subcaption}
\usepackage[preprint]{corl_2025} 

\title{Object-Centric Mobile Manipulation through SAM2-Guided Perception and Imitation Learning}

\author{Wang Zhicheng$^{1}$,  Satoshi Yagi$^{1}$, Satoshi Yamamori$^{1,2}$, Jun Morimoto$^{1,2}$
\thanks{$^{1}$Learning Machines Group, Graduate School of Informatics, Kyoto University, Kyoto, Japan.
        {\tt\small\{wang.zhicheng\}@lm.sys.i.kyoto-u.ac.jp,
        \{yagi, morimoto\}@i.kyoto-u.ac.jp
}}%
\thanks{$^{2}$Dept. of Brain Robot Interface, Computational Neuroscience Labs, ATR, Kyoto, Japan.
        {\tt\small yamamori@atr.jp
}}%
}

\begin{document}
\maketitle


\begin{abstract}
Imitation learning for mobile manipulation is a key challenge in the field of robotic manipulation. However, current mobile manipulation frameworks typically decouple navigation and manipulation, executing manipulation only after reaching a certain location. This can lead to performance degradation when navigation is imprecise, especially due to misalignment in approach angles. To enable a mobile manipulator to perform the same task from diverse orientations, an essential capability for building general-purpose robotic models, we propose an object-centric method based on SAM2, a foundation model towards solving promptable visual
segmentation in images, which incorporates manipulation orientation information into our model. Our approach enables consistent understanding of the same task from different orientations. We deploy the model on a custom-built mobile manipulator and evaluate it on a pick-and-place task under varied orientation angles. Compared to Action Chunking Transformer, our model maintains superior generalization when trained with demonstrations from varied approach angles. This work significantly enhances the generalization and robustness of imitation learning-based mobile manipulation systems.

\end{abstract}



\section{Introduction}
Manipulation tasks are a key milestone toward integrating robots into everyday life. They are particularly challenging because they require direct interaction with objects, demanding higher levels of precision and robustness \cite{rt1, rt2}.

Mobile manipulation -- the integration of navigation and object manipulation -- is essential for a domestic service robot, as it enables a single platform to perform diverse tasks in unstructured home environments \cite{hsr}. While fixed robots excel at repetitive factory operations, they lack the flexibility required to handle household chores. Consequently, the ability to generalize across tasks and environments becomes a critical capability for a truly versatile domestic assistant.

In recent years, learning-based control paradigms have gained significant traction for adapting to the unstructured environments encountered in manipulation tasks \cite{Aloha, diff}. Among these approaches, vision-based methods are especially appealing, as they leverage the ease of obtaining visual data and enable end-to-end mapping from raw images to action outputs. These methods—directly generating robot joint motions from camera images—facilitate the rapid deployment of domestic robots by eliminating the need for costly, large-scale sensor setups.

However, end-to-end mapping has several drawbacks. Since the agent merely reacts to raw RGB inputs without any explicit understanding of the objects or tasks in the image inputs, the resulting imitation policy often generalizes poorly. Therefore, many previous studies have conducted learning and inference for evaluation with the robot kept in a fixed position \cite{Aloha, diff, rt1, rt2, rtx}.


\begin{figure}[t]
  \centering
  \includegraphics[width=1\textwidth]{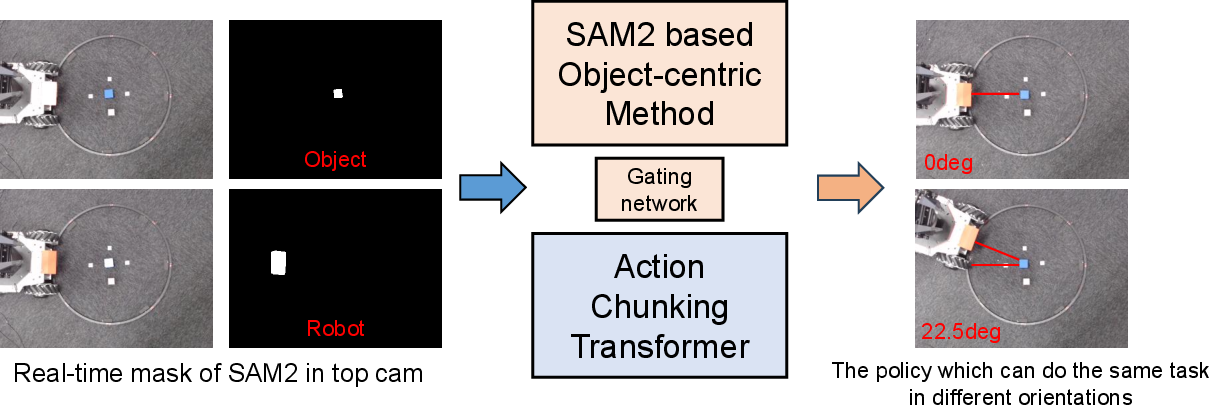}
  \caption{The overview of the method: first, RGB cameras capture real-time scene images. Next, we apply SAM2 to extract pixel-level masks for both the target object and the front edge of the robot. These masks are then embedded within action-chunking transformer through a gating network, enabling the robot to perform the same task across varying orientations.}
  \label{fig:overview}
\end{figure}

Rather than judging algorithms solely by success rates on fixed tasks, we believe it is crucial to carefully consider the challenges that arise when domestic robots perform mobile manipulation in real-world settings—where a far greater variety of situations can occur.
Indeed, in moving from simple, fixed tasks to a generalist robot model capable of handling multi-task, there remain key challenges at intermediate stages that the robot must overcome. For example, it must execute the same task reliably across varying environments, adapt to shifts in object positions, and cope with changes in orientation.

In testing mobile manipulation algorithms, we observed that decomposing navigation and manipulation in demonstrations causes the robot to simply replay the same motions—and yet vision‐based navigation often lacks the precision needed to reach an exact pre-grasp pose \cite{ishida, kazhoyan2020learning}. Small deviations in both distance and orientation between the robot and object then compound, creating what we term the ``orientation problem'' of mobile manipulation. As a result, these accumulated errors lead to a marked drop in operational performance compared to a fixed‐policy setup, where the robot executes manipulation under consistent, pre-defined conditions. Robust mobile manipulation therefore requires precise pose estimation and integrated error-compensation strategies that can handle the variability inherent in real‐world environments.

The purpose of applying an end-to-end policy to mobile manipulation is to bridge the gap between navigation and manipulation—enabling a single policy to generalize across different object orientations. This directly addresses the ``orientation problem," ensuring that the robot can robustly handle the same task even when its approach angle or relative pose varies.

Although end-to-end approaches are not perfect, they offer a higher performance ceiling than language-model-based control algorithms, due to their lightweight and fast characteristics. 
We think that, the restriction of the end-to-end policy is not the problem of the learning algorithm, but the problem of the data preprocessing and the action postprocessing. Therefore, we decided to use object-centric method, which can enhance the data preprocessing and action postprocessing to enhance the generalization ability of end-to-end models.
Object‐centric methods originated in computer vision as a way to isolate and describe individual objects within an image by using a pretrained model. Subsequently, the robotics community recognized that these object-level priors offer significant advantages for manipulation action control. Several studies have employed object-centric approaches within imitation-learning frameworks for fixed manipulation tasks, achieving significant performance gains \cite{viola, florence2019self, sieb2020graph}. 

Regarding specific pre-trained models, powerful pre-trained models such as SAM2 \cite{sam2} and RPN \cite{rpn} now enable rich object‐level interaction in images and video. Object‐centric representations serve as a strong prior, allowing agents to reason about an object’s position and characteristics rather than simply reacting to raw RGB inputs. Although these methods have already significantly boosted performance in fixed manipulation tasks, they have not yet been applied to mobile manipulation, where the robot’s base and the object continually shift relative to one another. In this context, an object‐centric approach naturally extends to an ``object–robot–centric'' framework that jointly encodes both object and robot poses.

An overview of our method is shown in Figure \ref{fig:overview}. We therefore leverage SAM2 as our segmentation-based prior: it produces accurate pixel-level masks without expensive fine-tuning and offers a promptable, unified detection-segmentation pipeline that remains robust under occlusion, lighting changes, and clutter. By incorporating this mask-based information, our end-to-end policy can better overcome orientation-induced errors and generalize across diverse real-world scenarios.

In the experiment, we demonstrated that the proposed SAM2-based object-centric approach improves both in-domain robustness and out-of-domain generalization in real-world environments. Specifically, we collected data for a stacking task by teleoperating an arm robot mounted on a mobile platform, lifting and placing boxes from two different angles. A neural network was then trained using this data. For evaluation, in addition to the two angles used during data collection (in-domain), we introduced a novel, unseen angle (out-of-domain) and performed the same task. As a result, our proposed method succeeded in all trials, revealing two key findings. First, it improved the success rate under in-domain conditions compared to the baseline method. Second, by leveraging in-domain knowledge, the proposed method was able to generalize to the out-of-domain task, where the baseline method failed.




\section{Related Works}
\textbf{Imitation learning of manipulation:} 
Several studies have demonstrated outstanding performance on manipulation tasks within fixed settings \cite{Aloha}\cite{diff}. On the method to increase the policy robustness, a work tries to fix the attach force control by using the whole body control \cite{Force}. In the field of increasing the generalization. In the other work, they try to create a robot general model \cite{rtx}.

Some works focus on the multi-view \cite{multiv}. Full multi-model robotic system \cite{big}. Some work use Vision language model to make a general policy, but need lots of data \cite{pi0} and inference time will be very low.
Some work want to use Gaussian splatting to build a realistic world model to increase the robustness of the robot manipulation in different object distribution \cite{dreamtomanipulate}. Some work want to use gaussian model to detect the position of the target object, in order to complete the tasks between subscenes \cite{tamma}. But it actually uses ground truth to get into the model, so it has some restrictions. Previous works also use LLM model to accomplish the multi-scene task \cite{saycan2022arxiv}. However, since their action planning often being in the form of policies, their generalization ability to scenes is relatively weak. There is also one work which employs a state-machine-like deep reinforcement learning model to enable switching between multi-task \cite{Science}.

Large language models deliver impressive policies but often demand massive datasets and suffer from high latency. For example, $\pi0$ is pre-trained on over 10,000 hours of dexterous robot demonstrations across seven distinct platforms and 68 tasks, incurring substantial data collection and GPU compute costs \cite{pi0}. At the same time, many fine-tuned large language model approaches still employ autoregressive action tokenization at only 3–5 Hz—far below the sub-50 ms inference budgets required for smooth, real-time control \cite{kim2025fine}. Compared to the large language model, by using an end-to-end policy, we can avoid requiring such large datasets and maintain a high control frequency thanks to fast inference speed.

Mobile aloha \cite{mobilealoha} employs an end-to-end policy, collecting data by first navigating to a target region and then initiating manipulation. Here, since they also employ whole-body control, the "manipulation" here represents actions that combine the manipulator arm’s actions with those of the mobile base. Due to the compound error \cite{compound}, in many mobile manipulation tasks, the navigation and manipulation stages are actually decomposed. Because the robot’s policy is trained solely on state and action pairs from demonstrations, it simply navigates to the demonstrated pose before starting its manipulation routine. Their algorithm achieves excellent success rates in single‐task scenarios, but its multi‐task capabilities have yet to be explored. To be clarified, the multi-task concept here does not merely mean using the same model for entirely different tasks—such as performing both pick-and-place and assembly on the same object—but rather the ability to execute the same manipulation task at varying relative positions within a workspace. 

\textbf{Object-centric method:}
In robotic manipulation, object‐level abstractions like poses or bounding boxes \cite{rpn, box} offer only coarse localization, omitting the rich shape and boundary details essential for precise interactions. Image segmentation fills this gap by assigning semantic labels to each pixel, enabling accurate contact‐point planning and on‐the‐fly re‐planning as objects or viewpoints shift. The Segment Anything Model 2 (SAM2) \cite{sam2} takes this further: its promptable interface, pixel-level masks zero-shot—no fine-tuning required—while its transformer backbone maintains robustness under occlusion, lighting changes, and clutter. With real-time inference speeds, SAM2 seamlessly integrates into mobile manipulation pipelines, enriching policy inputs with high-fidelity object representations and dramatically improving task understanding, orientation resilience, and generalization in unstructured home environments. Based on these advantages of SAM2, we decided to build an object-centric method on top of it to enhance the robustness and generalization of the end-to-end policy.


\section{Method}
\begin{figure}[ht]
  \centering
  \includegraphics[width=1\textwidth]{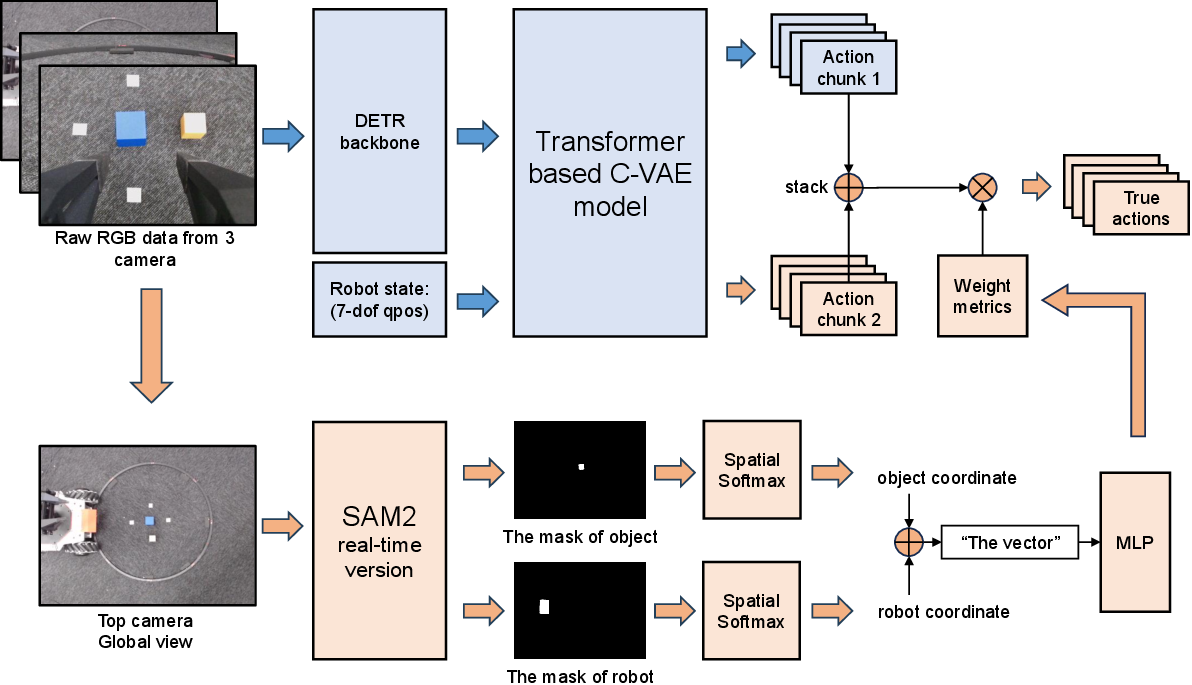}
  \caption{Model stucture: the blue part shows original action chunking transformer and the orange part shows our method}
  \label{fig:architecture}
\end{figure}

\subsection{Base model}

For the base model, we use action-chunking transformer presented by \cite{Aloha} and \cite{mobilealoha}. The main structure is a transformer-based conditional Conditional Variational Autoencoder (C-VAE) model, and the action chunking here means the output of the model is an action sequence rather than one action one time step. The encoder is the transformer encoder and decoder is the transformer encoder and decoder. As the paper has said, the latent representation is generated by encoder by position embedded action sequence and joint position. The input of the decoder part is the latent state, the joint position and RGB image information processed by the DETR backbone. The output is the manipulator’s action sequence. The baseline model achieves high levels of robustness and accuracy but suffers from limited generalization. According to our tests, on a single dataset, even slight deviations in angle or small object displacements cause the task to fail. Based on this characteristic, we aim to enhance the model’s generalization capability.

\subsection{SAM2-based Object-centric method}

We use the SAM2 to give the image from the top camera a real-time mask by assigning a point where the object is in the first frame. A mask from SAM2 is a binary image that highlights the exact pixels belonging to a specific object, enabling precise segmentation for downstream tasks. When using 2 frames tracking process separately, to have an object binary image and a robot binary image. Then we use spatial softmax to the binary mask, to make an object-centric coordinate and a robot-centric coordinate. We then get a ``vector'' point from the robot to the object. We want to use this vector to make the policy aware of the orientation change and make task-specific adjustments to the generation of action sequences. Therefore, we have some adjustments in the output of the original ACT structure, to add a gate network.

A gate network is a small neural module that takes context as input and produces a set of gating coefficients that dynamically weight among multiple sub-networks. In effect, it learns when and how much to rely on each expert’s output, enabling a conditional mixture-of-experts behavior in models.
In our approach, we feed the C-VAE’s hidden-layer outputs into an MLP for action sequences and employ a multi-head architecture to generate multiple sets of action sequences. We use the "vector" in the SAM2 part to generate a weight matrix by using a 2-layer MLP. We use the weight matrix to multiply the stack of the multi-head of the action sequence to get the true action sequence.

We mention that, due to the low inference speed of the SAM2 real-time version(max-speed around 7 FPS), we just use SAM2 first 10 time steps to call the SAM2 model. Then assign the following vector as the average of the values corresponding to these ten time steps. Because, in theory, the absolute position between the robot and the blue cube remains unchanged, which justifies the rationale for this operation. With that operation, we can make the control frequency of the model 30 FPS to make a smoother manipulation and compare with the original ACT policy more conveniently.



\begin{figure}[htbp]
    \centering
    \begin{subfigure}{0.4\textwidth}
        \centering
        \includegraphics[height=60mm]{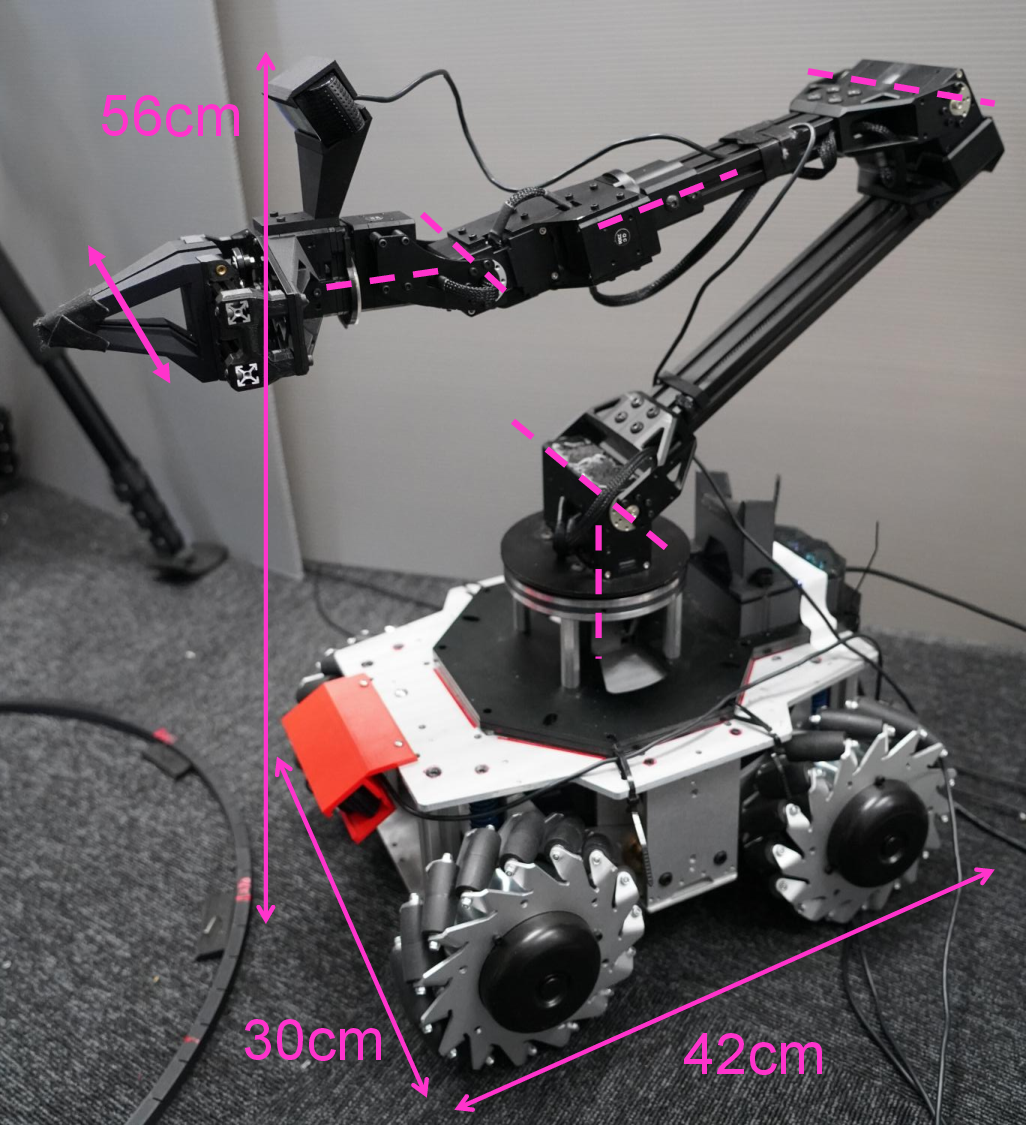}
        \caption{Robot}
    \end{subfigure}
    \hspace{5mm}
    \begin{subfigure}{0.5\textwidth}
        \centering
        \includegraphics[height=60mm]{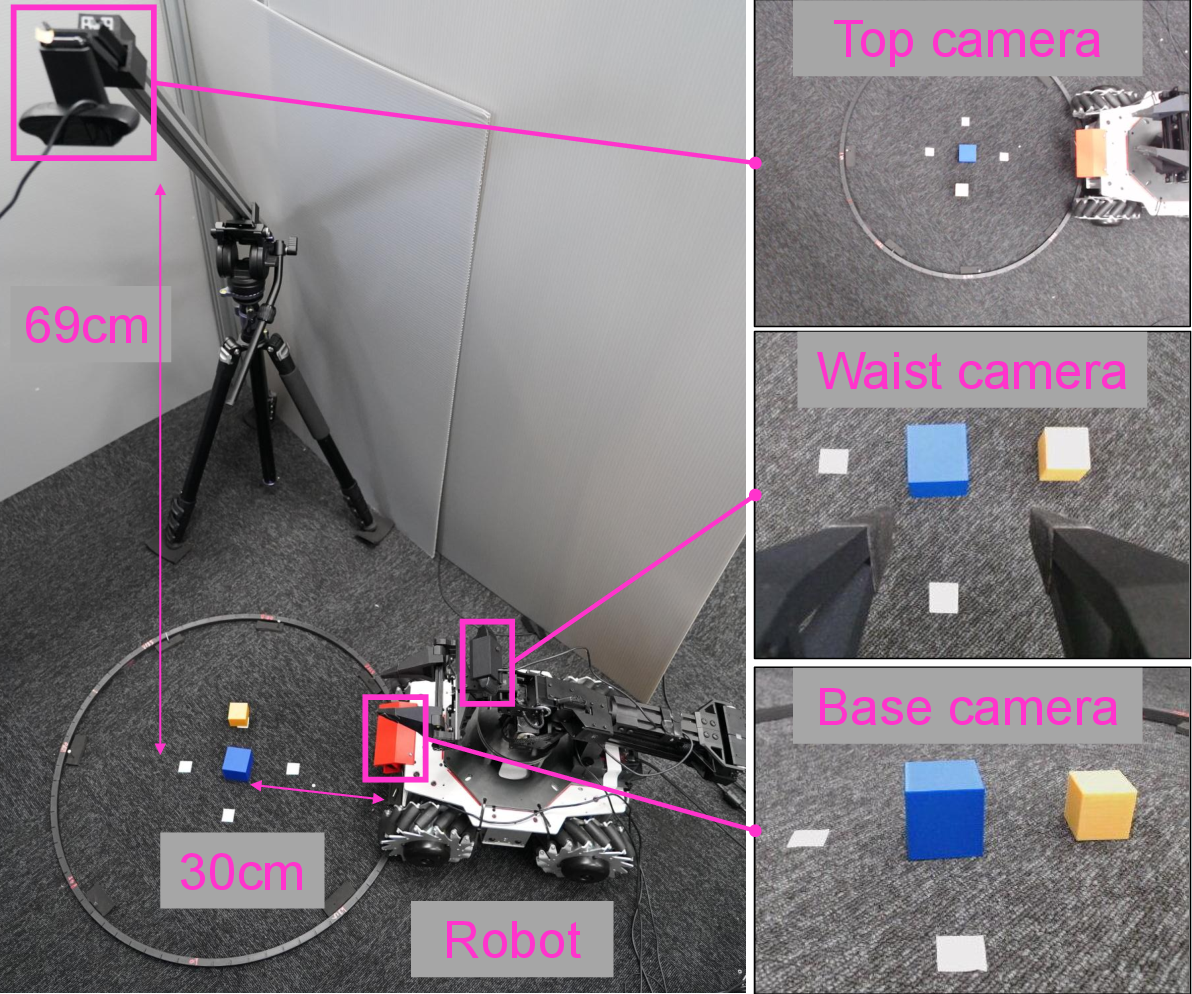}
        \caption{Experiment platform}
    \end{subfigure}
    \caption{(a) Robot, (b) Experiment platform}
    \label{fig:experiment}
\end{figure}

\subsection{Experient platform}
We build a novel mobile robotics research platform to do the orientation type research, with the manipulation part comes from Fixed Aloha and the mobile base with mecanum wheel.
About the data collection part, we mentioned that for the imitation learning research, the difficulties also happen in the data collection part, because typically, the researcher needs to collect more than 20 series data for one task setting and environment setting. Therefore, the usability is becoming important for the robot learning research platforms. 
In our study, we used a space mouse to control the movement and rotation of the mobile base as an intuitive interface for the operator. And by using the combination of the ROS nodes, we have completed the collaborative control of the master arm of aloha and the space mouse for precise, smooth, and labor-saving data collection. In our experimental setup, we assume the robot has completed the navigation phase, so the algorithm only covers the manipulation component. It should be clarified that, during experiments, we assume the robot has completed the navigation phase, so the algorithm only covers the manipulation component.

We design an experiment platform which can show the orientation problems. To control the parameters, we use a 3D-painting circle ring with a radius of 30 cm to restrict the distance from the robot to the object. There are some degree scales on the ring so we can make sure of the relative and abstract orientation degree of the mobile manipulation. There are 3 cameras as the source of image observation: the waist camera, the mobile base camera, and the top camera as shown in Figure 5. We mention that the base camera and the waist camera are moved with the mobile manipulator. The top camera is mounted on a fixed bracket above the scene to achieve a global observation. By pressing the wheels against the outer edge of the circular ring, the mobile base’s orientation remains fixed on the platform’s exact center. 

\subsection{Experiment design}

We use 2 models for training and evaluation, the baseline(Action Chunking Transformer) and our method.

We collect 30 demonstrations from 45 degree and 30 demonstrations from 0 degree and enhance training uniformity by employing data cross-shuffling upon these 60 demonstrations. We seperate it to 58 demonstrations to train and 2 demonstrations(one from 0 and one from 45) to test. After traning, we evaluate them on both 45 degree and 0 degree(in-domain), and 22.5 degree(out-of-domain).

The task is a stacking task in which the yellow cube is placed on top of the blue cube. Regarding the positions of the cubes, for every orientation setting of the mobile manipulator, their absolute positions remain the same: the blue cube is located at the exact center of the circular ring, while the yellow cube is placed slightly farther out at the 180-degree position. The overall task consists of two sub-tasks: first, reaching and grasping the yellow cube; and second, moving it above the blue cube. Because the relative position between the robot and the yellow cube changes with different orientation angles, the required action sequence differs significantly across orientation settings. 

\subsection{Experiment result} 
\begin{figure}[tbp]
\centering
\includegraphics[width=1.0\linewidth]{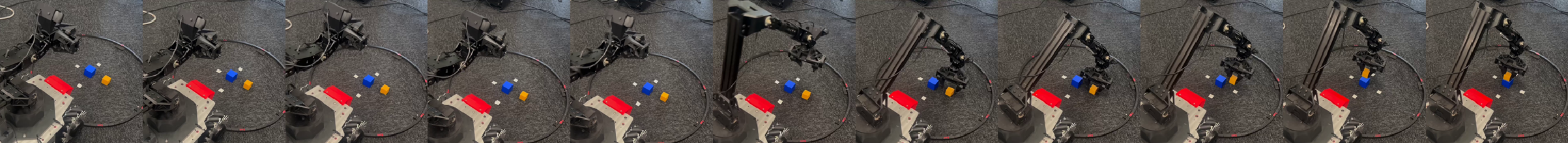}
(a) Proposed method under 0 deg condition (in-domain)\\
\vspace{5mm}
\centering
\includegraphics[width=1.0\linewidth]{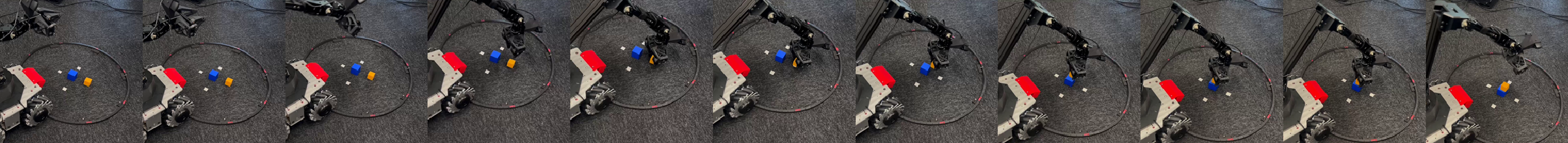}
(b) Proposed method under 22.5 deg condition (out-of-domain)\\
\centering
\includegraphics[width=1.0\linewidth]{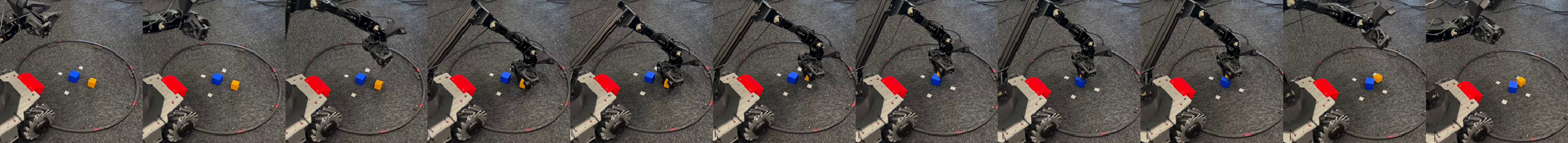}
\vspace{5mm}
(c) ACT under 22.5 deg condition (out-of-domain)\\
\caption{Representative results: (a) Proposed method under the 0 deg condition (in-domain), (b) Proposed method under the 22.5 deg condition (out-of-domain), and (c) Action Chunking Transformer under the 22.5 deg condition (out-of-domain). For each condition, the robot's motion during the first 10 seconds from the start is illustrated from left to right at 1-second intervals. (a) and (b) demonstrate that the proposed method successfully places the yellow cube on top of the blue cube, whereas the baseline method (c) fails to achieve this.}
\label{fig:result}
\end{figure}
We collect 30 demonstrations from 45 degree and 30 demonstrations from 0 degree and enhance training uniformity by employing data cross-shuffling upon these 60 demonstrations. When then train the baseline policy and our method on these demonstrations, then evaluate them on both 45 degree and 0 degree(in-domain), and 22.5 degree(out-of-domain).
We record in in-domain performance, the success rate of baseline policy is 90\% in 45 degree and 70\% in 0 degree, and the success rate of our method is 100\% in both 45 degree and 0 degree, which demonstrates our method’s enhancement of multi-task capabilities.
For the out-of-domain setting in 45 degree, the success rate of the baseline is 20\% and our method is 100\%, which demonstrates a substantial enhancement in generalization capability. The results are presented in Table 1.

\captionsetup[table]{skip=10pt}
\title{Result camparision}
\begin{table}
  \caption{Result camparision}
  \centering
  \begin{threeparttable}
 
    \begin{tabular}{ccc@{\qquad}c}
      Model & 0 degree & 45 degree & 22.5 degree \\ \midrule\midrule 
        \makecell{Action Chunking Transformer} & 70\% & 90\% & 20\% \\
     \cmidrule(l r){1-4}
      Our Method &  100\% &  100\% & 100\%\\ \midrule\midrule
    \end{tabular}

\end{threeparttable}
\end{table}

\section{Discussion}
\label{sec:discussion}
From our experimental results, we demonstrate that incorporating our object-centric method significantly improves both in-domain and out-of-domain performance compared to the original Action Chunking Transformer. We attribute this success to our multi-head action architecture and the learned weight matrix, which together enable dynamic task adaptation. Specifically, each head proposes a candidate action sequence, and the weight matrix selects and combines these proposals based on the embedded orientation vector. During training, varying object–robot relative poses, as captured by the SAM2-derived masks, produce distinct orientation embeddings; consequently, the weight matrix learns to emphasize the appropriate action head for each pose. This mechanism enables the policy to generalize across varying orientations by dynamically selecting and combining the most appropriate action head for each specific pose. In summary, we leverage higher-dimensional information to assist the policy in generating action strategies under the orientation problem.

\section{Conclusion}
\label{sec:conclusion}
In this work, we implement a SAM2-based object-centric method that significantly enhances the in-domain robustness and out-of-domain generalization of end-to-end models. This capability can be used to address the orientation problem, thereby improving the efficiency and robustness of mobile manipulators when performing the same task in diverse environments. The main point of this paper is how can we handle the middle state by using the data from left and right sides demonstrations. 


\clearpage
\acknowledgments{} 


\section*{Limitations} 
Our approach has two notable limitations. First, integrating the SAM2 segmentation model adds significant computational overhead, increasing inference latency and potentially reducing the control loop frequency. Second, relying on a fixed, ceiling-mounted global camera restricts the robot’s operational workspace: if the mobile base travels beyond the camera’s field of view during navigation, the segmentation and orientation-embedding pipeline cannot function reliably. 
\bibliography{example}  

\end{document}